# Extraction of domain-specific bilingual lexicon from comparable corpora: compositional translation and ranking

*Estelle DELPECH[1], Béatrice DAILLE[1], Emmanuel MORIN[1], Claire LEMAIRE[2,3]*
(1) UNIVERSITÉ DE NANTES – LINA UMR 6241, 2 rue de la Houssinière, BP 92208, 44322 Nantes, Cedex 3, France
(2) UNIVERSITÉ STENDHAL – GRENOBLE 3, BP 25, 38040 Grenoble Cedex 9, France
(3) LINGUA ET MACHINA, c/o Inria Rocquencourt BP 105, Le Chesnay Cedex 78153, France
(1){name.surname}@univ-nantes.fr

(2){initials}@lingua-et-machina.com

ABSTRACT

This paper proposes a method for extracting translations of morphologically constructed terms from comparable corpora. The method is based on compositional translation and exploits translation equivalences at the morpheme-level, which allows for the generation of "fertile" translations (translation pairs in which the target term has more words than the source term). Ranking methods relying on corpus-based and translation-based features are used to select the best candidate translation. We obtain an average precision of 91% on the Top1 candidate translation. The method was tested on two language pairs (English-French and English-German) and with a small specialized comparable corpora (400k words per language).

TITLE AND ABSTRACT IN ANOTHER LANGUAGE, FRENCH

## Extraction de lexiques bilingues spécialisés à partir de corpus comparales : traduction compositionnelle et ordonnancement

Cet article propose une méthode permettant d'extraire des traductions de termes morphologiquement construits à partir de corpus comparables. La méthode se base sur la traduction compositionnelle et exploite des équivalences traductionnelles au niveau morphologique, ce qui nous permet de générer des traductions "fertiles" (des paires de traductions dans lesquelles le terme cible a plus de mots que le terme source). Des méthodes d'ordonnancement s'appuyant sur des traits extraits du corpus et des paires de traduction sont utilisées pour sélectionnner la meilleur traduction candidate. Nous obtenons une précision de 91% sur le Top1 en moyenne. La méthode a été testée sur deux paires de langues (anglais-français et anglais-allemand) et sur un corpus comparable spécialisé de petite taille (400k mots par langue).

MOTS-CLÉS : TRADUCTION ASSISTÉE PAR ORDINATEUR, TRADUCTION AUTOMATIQUE, CORPUS COMPARABLES, LEARNING-TO-RANK, COMPOSITIONNALITÉ, TERMINOLOGIE

## Introduction

Comparable corpora are composed of texts in different languages which are not translations but deal with the same subject matter and were produced in similar situations of communication. They are used in Computer-Aided Translation to provide technical translators with domain-specific bilingual lexicons when there is no parallel data available (e.g. translation memories, multilingual terminologies). This situation happens when translators have to translate texts which deal with emerging technical domains or when the translation is done from/to an under-resourced language. Comparable corpora also have the advantage of containing more idiomatic expressions than parallel corpora do because the target texts do not bear the influence of the source language. Indeed, Baker (1996) observed that translated texts tend to bear features like explicitation, simplification, normalization and levelling out. As a consequence, one of the difficulties with comparable corpora is that the translation of a source term may not be present in its "normalized" or "canonical" form but rather in the form of a morphological or paraphrastic variant (e.g. *post-menopausal* translates to *après la ménopause* 'after the menopause' instead of *post-ménopausique*). Another limitation is that algorithms output, for each source term, a set of candidate translations instead of just one target term. This state of affairs makes it very challenging for translators to use lexicons extracted from comparable corpora in real-life situations (Delpech, 2011).

The solution that consists in increasing the size of the corpus in order to find more translation pairs or to extract parallel segments of text (Fung & Cheung, 2004; Rauf & Schwenk, 2009) is only possible when large amounts of texts are available. In the case of the extraction of *domain-specific* lexicons, we quickly face the problem of data scarcity: in order to extract high-quality lexicons, the corpus must contain text dealing with very specific subject domains and the target and source texts must be highly comparable. If one tries to increase the size of the corpus, one takes the risk of decreasing its quality by adding out-of-domain texts. Studies support the idea that the quality of the corpora is more important than its size. Morin *et al.* (2007) show that the discourse categorization of the documents increases the precision of the lexicon despite the data sparsity. Bo & Gaussier (2010) show that they improve the quality of the extracted lexicon if they improve the comparability of the corpus by selecting a smaller – but more comparable – corpus from an initial set of documents.

This paper proposes methods for ranking and extracting canonical translations as well as translation variants, with a special focus on the extraction of *fertile* translations. In parallel texts processing, the notion of fertility has been defined by Brown *et al.* (1993). They defined the fertility of a source word *e* as the number of target words to which *e* is connected in a randomly selected alignment. Similarly, we call a fertile translation a translation pair in which the target term has more words than the source term. The identification of fertile translations is useful because (i) they frequentlty correspond to non-canonical translations, e.g. paraphrastic variants and (ii) they tend to correspond to vulgarized forms of technical terms (e.g. « cytotoxic » vs. « toxic to the cells ») which are useful when the translator translates lay science texts. Up to now, fertility has received little attention in the field of comparable corpora processing. To our knowledge, only Daille & Morin (2005) and Weller *et al.* (2011) tried to extract translation pairs of different lengths from comparable corpora. Daille & Morin (2005) focus on the specific case of multi-word terms whose meaning is not compositional and tried to align these multi-word terms with either single-word terms or multi-word terms using a context-based approach. Weller

*et al.* (2011) concentrate on translating noun compounds to noun phrases. Similar to the approach presented here, Claveau & Kijak (2011) use translation equivalences between morphemes to generate translations and can handle fertility. However it is not suited for comparable corpora since it requires domain-specific parallel data (in their case, a multilingual terminology) to learn alignment probabilities.

Our method is based on compositional translation. We chose this approach because: (i) according to Namer & Baud (2007), compositional terms form a major part of the new terms found in technical and scientific domains, this is not restricted to the field of biomedicine as it is generally believed ; (ii) compositionality-based methods have been shown to clearly outperform context-based ones for the translation of terms with compositional meaning, both in terms of translation accuracy and rank of the correct candidate translation (Morin & Daille, 2010) ; (iii) we believe that compositionality-based methods offer the opportunity to generate fertile translations if combined with a morphology-based approach. This method, which we call morpho-compositional translation, consists in: **(i) decomposing** the source term into morphemes: *post-menopause* is split into *post-* + *menopause*[1] ; **(ii) translating** the morphemes to bound morphemes or fully autonomous words: *post-* becomes *post-* or *après*, *menopause* becomes *ménopause* ; **(iii) recomposing** the translated elements into a target term: *post-ménopause* 'post-menopause', *après la ménopause* 'after the menopause'. Fertile translations can be generated because we allow bound morphemes to be translated to autonomous lexical items (e.g. prefix *post-* → preposition *après*). The proposed ranking methods exploit various corpus-based and translation-based features.

This paper falls into 4 sections. Section 1 outlines recent research in compositional approaches to bilingual lexicon extraction. Section 2 explains the methods we designed for translation generation and ranking. Section 3 describes our experimental data. Section 4 presents and discusses the results of our experimentations.

# 1    Compositional approaches to bilingual lexicon extraction

The core of compositional translation consists in generating candidate translations following the principle of compositionality: "*the meaning of the whole is a function of the meaning of the parts*" (Keenan & Faltz, 1985, pp. 24-25). Once the candidate translations have been generated, one generally ranks them and selects the TopN candidate translations. Generation methods are described in section 1.1. Ranking methods are described in section 2.3.

## 1.1    Generation methods

Compositional translation consists in decomposing the source term into atomic components, translating these components into the target language and recomposing the translated components into target terms. Existing implementations differ on the kind of atomic components they use for translation.

**Lexical compositional translation** (Baldwin & Tanaka, 2004; Grefenstette, 1999; Morin & Daille, 2009; Robitaille *et al.*, 2006) deals with multi-word term to multi-word term alignment and uses lexical words as atomic components: *rate of evaporation* is translated into French as *taux d'évaporation* by translating *rate* to *taux* and *evaporation* to *évaporation* using dictionary lookup. Recomposition may be done by permuting the translated components (Morin & Daille, 2010) or with translation patterns (Baldwin & Tanaka, 2004).

---
[1]We use the following notations: trailing hyphen for prefixes (*a-*), leading hyphen for suffixes (*-a*), both for confixes (*-a-*), no hyphen for autonomous morphemes (*a*) and a plus sign (+) for intra-word morpheme boundaries. The term *confix* is borrowed from (Martinet, 1979) and refers to neoclassical (Latin or Ancient Greek) roots.

**Sublexical compositional translation** deals with single-word term translation. The atomic components are subparts of the source single-word term. Cartoni (2009) translates neologisms created by prefixation with a formalism called Bilingual Lexeme Formation Rules. Atomic components are the prefix and the lexical base: Italian neologism *ricostruire 'rebuild'* is translated into French *reconstruire* by translating the prefix *ri-* to *re-* and the lexical base *costruire* as *construire*. Weller *et al.* (2011) translate two types of single-word term. German single-word terms formed by the concatenation of two neoclassical roots are decomposed into these two roots, then the roots are translated into target language roots and recomposed into an English or French single-word term, e.g. *Kalori$_1$metrie$_2$* is translated as *calori$_1$metry$_2$*. German Noun$_1$+Noun$_2$ compounds are translated into French and English Noun$_1$ Noun$_2$ or Noun$_1$ Prep Noun$_2$ multi-word terms, e.g. Elektronen$_{N1}$-mikroskop$_{N2}$ is translated to *electron$_{N1}$ microscope$_{N2}$*. Garera & Yarowsky (2008) translate various compound sequences (Noun$_1$+Noun$_2$, Adj$_1$+Noun$_2$ ...). They generate an English literal gloss of the compounds with the compositional method (for instance, the English gloss for the Albanian word *hekurudhë 'railway'* is *iron path*). Then, they search for entries in *Lx*-to-English dictionaries where the entry in language *Lx* is a word-to-word translation of the English gloss (e.g. *iron path* matches the German entry *Eisenbahn* and the Italian entry *ferrovia*). The final candidate translations are the fluent English translations proposed by the bilingual dictionaries (e.g. *Eisenbahn* and *ferrovia* both translate to *railway* ; *railway* is considered as a potential translation for *hekurudhë*).

## 1.2 Ranking and selection methods

Generally, compositional translation generates several possible translations for one source term. One has to find a way to rank the translations from the most to the least reliable. Garera & Yarowsky (2008) tried two ranking methods: (i) a probability score $\mathcal{P}$ based on the number of different languages exhibiting the association between the literal gloss and the fluent translation ; (ii) the probability score $\mathcal{P}$ combined with the similarity of the source and target words' contexts using context-based methods like in the work of Rapp (1995) and Fung (1997). Robitaille *et al.* (2006) extract translation pairs from a corpus built by querying a search engine with a set of seed translation pairs. They select the candidate translations which are semantically related to the target seed terms. The semantic similarity measure is based on the number of hits containing the seed term and/or the candidate translation (Jaccard coefficient). Other works simply select the candidate translations which occur in the target corpus (Weller *et al.*, 2001 ; Morin and Daille, 2010) or which are significantly attested on the Web (Cartoni, 2009).

Only Baldwin and Takana (2004) use machine learning. They train a SVM classifier with corpus-based, dictionary-based and translation pattern-based features and use the value returned by the classifier (a continuous value between -1 and +1) to rank the candidate translations. Their approach is tantamount to point-wise approaches in learning-to-rank. To our knowledge, no research work has investigated the possible contribution of advanced learning-to-rank algorithms to candidate translations ranking. Learning-to-rank algorithms are widely used in Information Retrieval for ranking documents from the most to the least relevant to a given query (Li, 2011). They can be easily ported to the problem of ranking the candidate translations of a source term.

There exists three families of learning-to-rank algorithms: **point-wise** (for a given a query-document pair, predict its relevance score or label: ranking is treated as a regression or classification problem), **pair-wise** (for a given query and two documents, indicate which of the two documents is the most relevant) and **list-wise** (given a query and a list of documents, indicate how to order the documents: this last family straightforwardly represents learning-to-rank problem). According to the tests of Liu (2009), list-wise algorithms generally outperform the two other approaches.

## 1.3   Challenges of compositional translation

Compositional translation faces four main challenges which are: **morphosyntactic variation:** source and target terms' morphosyntactic structures are different: *anti-cancer$_{NOUN}$* → *anti-cancéreux$_{ADJ}$ 'anti-cancerous'* ; **lexical variation:** source and target terms contain semantically related - but not equivalent - words: *machine translation* → *traduction automatique 'automatic translation'* ; **terminological variation:** a source term can be translated to different target terms: *oophorectomy* → *ovariectomie 'oophorectomy', ablation des ovaires 'removal of the ovaries'* ; **fertility:** the target term has more content words than the source term. Note that fertility can have two origins. In the case of *surface* **fertility**, the target term has more words than the source term but source and target terms have the same number of morphemes. Source and target languages differ in the way they concatenate morphemes to form words: *bi-dimensional* → *deux dimensions 'two dimensions'*. In the case of *semantic* **fertility,** the target term has more morphemes than the source term. Source and target languages differ in the way they combine elements of meaning to create new words: *voie de glace 'route of ice'* → *ice climbing route, aquarelle* (not decomposable) → *water color*.

Solutions to morphosyntactic, lexical and to some extent terminological variation have been proposed in the form of thesaurus lookup (Robitaille *et al.*, 2006), morphological derivation rules (Morin & Daille, 2010), morphological variant dictionaries (Cartoni, 2009) or morphosyntactic translation patterns (Baldwin & Tanaka, 2004; Weller *et al.*, 2011). Although it is not specifically outlined in their paper, the work of Garera & Yarowsky (2008) can theoretically handle semantic fertility and lexical divergence. However, their method depends on a large number of dictionaries with a substantial coverage of compounds. Surface fertility has been addressed by Weller *et al.* (2011) for the specific case of German N$_{OUN}$+N$_{OUN}$ compounds. The method presented here, which we call "morpho-compositional translation", is able to deal with surface fertility and to generate and rank translations for a large variety of morphologically constructed words.

## 2   Translation method

## 2.1   Principle of morpho-compositional translation

The idea of morpho-compositional translation is to apply the principle of compositional translation at the morpheme-level rather than at the lexical level and to allow translation equivalences between bound and autonomous morphemes in order to generate fertile translations. It relies on the assumptions that: **(i) a lexical item can be decomposed into smaller components**. These components may be **free**, i.e. they can occur in texts as autonomous lexical items like *toxicity* in *cardiotoxicity* or **bound**, i.e. they cannot occur as autonomous lexical items, in that case they correspond to bound morphemes like *-cardio-* in *cardiotoxicity* ; **(ii) a bound component can be translated to an autonomous or a bound component:** *-cardio-* can be translated to *-cardio-* or *cœur* 'heart' or *cardiaque* 'cardiac'. Thus, *cardiotoxicity* can be translated to *toxicité cardiaque* 'cardiac toxicity' or *toxicité pour le cœur* 'toxicity to the heart' or *cardiotoxicité* 'cardiotoxicity'.

Like other sublexical approaches, the main idea behind morpho-compositional translation is to go beyond the word level and work with subword components. In our case, these components are morpheme-like items which either (i) bear referential lexical meaning like confixes (*-cyto-, -bio-*) and autonomous lexical items (*cancer, toxicity*) or (ii) can substantially change the meaning of a word, especially prefixes (*anti-, post-*) and some suffixes (*-less, -like*). Unlike other approaches, morpho-compositional translation is not limited to small set of source-to-target structure equivalences. It takes as input a morphologically constructed single-word term which can be the result of prefixation '*pretreatment*', confixation '*densitometry*', suffixation '*childless*', compounding '*anastrozole-associated*' or any combinations of the four. Its output is a set of single or multi-word candidate translations. For instance, *postoophorectomy* may be translated to *postovariectomie* '*postoophorectomy*' or *après l'ovariectomie* 'after the oophorectomy' or *après l'ablation des ovaires* 'after the removal of the ovaries'.

Section 3.2 explains the algorithm for generating candidate translations. Section 3.3 describes different methods for ranking the candidate translations.

## 2.2 Generation algorithm

The generation method is described in the algorithm 1. A detailed version of the algorithm can be found in the feasibility study of Delpech *et al.* (2012).

---

**Algorithm 1** Generate translations

**Require:** *source_term, target_corpus*
*translations* ← Ø
**for all** $\{c_1, ... c_i\}$ in DECOMPOSE (*source_term*) **do**
   **for all** $\{e_1, ... e_j\}$ in CONCATENATE ($\{c_1, ... c_i\}$) **do**
     **for all** $\{t_1, ... t_k\}$ in {TRANSLATE ($e_1$) × … TRANSLATE ($e_j$)} **do**
       **if** $k \neq j$ **then**
         continue
       **for all** $\{t_1, ... t_k\}$ in PERMUTATE ($\{t_1, ... t_k\}$) **do**
         **for all** $\{w_1, ... w_l\}$ in CONCATENATE ($\{t_1, ... t_k\}$) **do**
           **for all** *match* in MATCH ($\{w_1, ... w_l\}$, *target_corpus*) **do**
              add *match* to *translations*
return *translations*

---

**The DECOMPOSE function** splits the source term into minimal components $\{c_1, ... c_i\}$ by matching substrings of the term with lists of prefixes, confixes, suffixes and lexical items and respecting some length constraints on the substrings. When several splittings are possible, only the ones with the highest number of components are retained.

**The CONCATENATE function** generates all possible concatenations of a list of components. For example, if the term "*abc*" has been split into 3 components *{a, b, c}*, then there are 4 different concatenations : *{a,bc}, {ab,c}, {a,b,c}, {abc}* (for *n* components, we have $2^{n-1}$ possible concatenations). When called used after the decomposition, the concatenation of the components increases the chances of matching the entries of the linguistic resources used by the TRANSLATE function. When called after the permutation, the concatenation is used to recreate a set of target words *{w₁, ...w₁}* from the set of translated components.

**The TRANSLATE function** uses two kinds of linguistic resources to generate translations. Bilingual resources map elements across languages. Variation resources are used to handle variation at the lexical and morphological level. Hence, the output of TRANSLATE(*e*) correspond to *Trans(e) U Trans(Var$^{src}$(e)) U Var$^{tgt}$(Trans(e))* where *Trans* is a bilingual resource, *Var* a variation resource, *src* is the source language and *tgt* is the target language. For example, *toxic* can be translated to *toxique 'toxic', toxicité 'toxicity'* or *vénéneux 'poisonous'*. If one element can not be translated then the translation of the whole fails.

**The PERMUTATE function** serves to capture the fact that components' order may be different in the source and target language (distortion). As a general rule, *O(n!)* procedures should be avoided but we are permuting small sets (up to 4 items).

**The MATCH function** returns a series of tokens which occur in the target corpus and whose lemmas match the generated target words *{w₁, ... w₁}*. We allow for 3 stop words between each lemma. For example, if the system generates the target words *{toxique, cellule} 'toxic, cell'* from *cytotoxic*, it will match *"toxique pour les cellules" 'toxic to the cells'*. We consider that two matches are one and the same translation if they correspond to the same series of *(lemma, part-of-speech)* pairs. For example, *toxique pour les cellules* and *toxique pour la cellule 'toxic to the cell'* correspond to the same translation.

## 2.3    Ranking methods

We have considered four parameters for ranking the translations: frequency of the translation, part-of-speech translation probability, context similarity and the reliability of the resources used for translating the components of the source term. These parameters can be used separately or in a combined manner.

**The frequency (FREQ)** corresponds to the number of occurrences of the translation in the target corpus divided by the total number of words in the target corpus.

**The part-of-speech translation probability (POS)** corresponds to *P(y|x)*, the probability that a source term with part-of-speech *x* will be translated to a target term with part(s)-of-speech *y*, e.g. it is more probable that a NOUN is translated by another NOUN or by a NOUN PREP NOUN sequence rather than an ADVERB. The part-of-speech translation probabilities were acquired by running the software ANYMALIGN (Lardilleux, 2008) on the EMEA corpus (Tiedemann, 2009) which had been previously pos-tagged with the linguistic analyzer XELDA[2]. ANYMALIGN outputs a phrase translation table. Each line of the translation table corresponds to an alignment *a = {lem$_s$, pos$_s$, lem$_t$, pos$_t$, p(s|t), p(t|s)}* where *pos$_s$* is the parts-of-speech of the source phrase, *pos$_t$* is the parts-of-speech of the target phrase and *pos(t|s)* is the probability of translating the source phrase to the target phrase. From these alignments, we obtain *P(y|x)* with the following formula:

$$P(y|x) = \frac{\sum_{\{a \in A | pos_s = x, pos_t = y\}} p(t|s)}{\sum_{\{a \in A | pos_s = x\}} p(t|s)}$$

---
[2]http://www.temis.com

**The context similarity (CONT)** corresponds to the method used for ranking translations in context-based approaches. For each source term and target term we build a *context vector*. This vector indicates the number of times the term co-occurs with each word of the corpus within a contextual window of 5 words around the term. The number of co-occurrences is normalized with the log-likelihood ratio (Dunning, 1993). Then, the vector of the source term is translated into the target language. Finally, the source vector and the target vector are compared: the most similar the vectors, the most likely the target and source terms are translations of each other. The similarity between source vector *s* and target vector *t* is computed with the weighted jaccard:

$$WeightedJaccard(s,t) = \frac{\sum_{w \in s \cap t} min(c(s,w), c(t,w))}{\sum_{w \in s \cap t} max(c(s,w), c(t,w)) + \sum_{w \in s \setminus t} c(s,w) + \sum_{w \in t \setminus s} c(t,w)}$$

where *c(s, w)*, respectively *c(t, w)*, is the normalized number of co-occurrences between the source, respectively target, term and word *w*. Note that for multi-word terms, the context vector corresponds to the union of the context vectors of the lexical words that compose the multi-word term.

**The resources score (RESO)** corresponds to:

$$RESO(t) = \frac{1}{|T|} \sum_{i=1}^{|T|} rel(T_i)$$

where *T* is the total number of components in the target term *t* and *rel(T$_i$)* is the reliability of the target component *T$_i$*. The reliability of a component is a float value between 0 and 1 inclusive. It depends on the nature of the component and on the resources which were used to generate it. We defined 8 types of target components: 1) the target component is a lexical item found in a general-language dictionary ; 2) target component is a lexical item which was found by cognate matching ; 3) target component is a lexical item which is a lexical variant of the translation (found in general language dictionary or by cognate matching) ; 4) target component is a lexical item which is a morphological variant of the translation ; 5) target component is a lexical item which is the translation of a bound morpheme ; 6) target component is a prefix ; 7) target component is a confix ; 8) target component is a suffix. We tuned the reliability values associated to these 8 types of target components empirically: we tested several arrangements[3] of reliability values on a training dataset (described in section 3.3) and retained the arrangement that gave the best rankings.

**Scores combination (COMBI)** is the linear combination of the FREQ, POS, CONT and RESO scores.

**Learning-to-rank algorithms (LTR)** were also tested. We tried three list-wise algorithms: AdaRank (Li & Xu, 2007), Coordinate Ascend (Metzler & Croft, 2000) and LambdaMart (Wu *et al.*, 2010). We used the implementations available in the RankLib software[4]. The predictive variables are the FREQ, POS, CONT and RESO scores. We trained the models on the training dataset described in section 3.3.

---

[3] all 8-arrangement with repetition of {0, 0.2, 0.4, 0.6, 0.8, 1}
[4] http://people.cs.umass.edu/~vdang/ranklib.html. We set the metric to optimize on the training data to MAP (Manning *et al.*, 2008) ; all other parameters were left to default.

## 3   Data

We worked with 3 languages: English as source language and French and German as target languages.

### 3.1   Comparable corpora

Our corpus is composed of specialized texts from the medical domain dealing with breast cancer. We define specialized texts as texts being produced by domain experts and directed towards either an expert or a non-expert readership (Bowker & Pearson, 2002). The texts were collected from scientific papers portals and from information websites targeted to breast cancer patients and their relatives. Each corpus has approximately 400k words (cf. table 1). All texts were pos-tagged and lemmatized with XELDA. We also computed the comparability of the corpora[5]. The English-French corpus' comparability is 0.71 and the English-German corpus' comparability is 0.45. The difference in comparability can be explained by the fact that German texts on breast cancer were hard to find (especially scientific papers): we had to collect texts in which breast cancer was not the main topic. This may have added out-of-domain words.

|                        | EN      | FR       | DE      |
|------------------------|---------|----------|---------|
| Expert readership      | 218.3k  | 267.2k   | 197.2k  |
| Non-expert readership  | 198.2k  | 184.5k   | 201.7k  |
| TOTAL                  | 416.5k  | 451.75k  | 398.9k  |

TABLE 1: Composition and size of corpora (nb. of words)

### 3.2   Resources for generation

Tables 2 and 3 show the size of the resources we used for generation.

**General language dictionary:** We used the dictionary which is part of the XELDA software. This dictionary was used for generating translations but also for computing the corpus comparability and for translating the context vectors for the context similarity measure (CONT score).

**Cognate dictionary:** We built this resource automatically by extracting pairs of cognates from the comparable corpora. We used the same technique as Hauer & Kondrak (2011): a SVM classifier trained on examples taken from online dictionaries[6].

**Morpheme translation table:** this resource was created manually by translators since there exists no publicly available morphology-based bilingual dictionary. This translation table links the English bound morphemes contained in the source terms to their French or German equivalents (which can be bound morphemes or lexical items).

In order to handle the variation phenomena described in section 1.3, we used a **dictionary of synonyms** and lists of **morphologically related words**. The dictionary of synonyms is part of the XELDA software. Morphologically related words were collected by stemming the words of the comparable corpora and the entries of the bilingual dictionary with the algorithm of Porter (1980).

---
[5]We used the measure defined by Bo & Gaussier (2010) which indicates, given a bilingual dictionary, the expectation of finding, for each word of the source corpus, its translation in the target corpus and *vice-versa*.
[6]http://www.dicts.info/uddl.php

The D<small>ECOMPOSE</small> function uses the entries of the morpheme translation table (242 entries) and a list of 85k lexical items composed of the entries of the general language dictionary and English words extracted from the Leipzig Corpus (Quasthoff *et al.*, 2006).

|  | EN → FR | EN → DE |
|---|---|---|
| General language | 38k → 60k | 38k → 70k |
| Domain specific | 6.7k → 6.7k | 6.4k → 6.4k |
| Morphemes (TOTAL) | 242 → 729 | 242 → 761 |
|   Prefixes | 50 → 134 | 50 → 166 |
|   Confixes | 185 → 574 | 185 → 563 |
|   Suffixes | 7 → 21 | 7 → 32 |

T<small>ABLE</small> 2: Nb. of entries in the multilingual resources

|  | EN → EN | FR → FR | DE → DE |
|---|---|---|---|
| Synonyms | 5.1k → 7.6k | 2.4k → 3.2k | 4.2k → 4.9k |
| Morphological families | 5.9k → 15k | 7.1k → 18k | 7.4k → 16k |

T<small>ABLE</small> 3: Nb. of entries in the monolingual resources

### 3.3 Datasets for evaluation and training

We extracted morphologically constructed source terms from the English texts in a semi-supervised manner: **(i)** we wrote a short seed list of English bound morphemes. We automatically extracted from the English texts all the words that contained these morphemes. For example, we extracted the words *postchemotherapy* and *poster* because they contained the string *post-* which corresponds to a bound morpheme of English ; **(ii)** The extracted words were sorted: those which were not morphologically constructed were eliminated (like *poster*), and those which were morphologically constructed were kept (like *postchemotherapy*). The morphologically constructed words were manually split into morphemes. For example, *postchemotherapy* was split into *post-*, *-chemo-* and *therapy* ; **(iii)** if some bound morphemes which were not in the initial seed list were found when we split the words during step (ii), we started the whole process again, using the new bound morphemes to extract new morphologically constructed words.

We also added hyphenated terms like *ER-positive* to our list of source terms. With this method, we collected 2025 source terms. Then, we excluded all the source terms which could be translated with the general language dictionary and whose translation was present in the target corpus. Finally for each language pair, we divided the source terms into two groups:

**The evaluation dataset** contains source terms which could be translated with the UMLS meta-thesaurus (Bodenreider, 2004) and whose translation was in the target corpus – these terms, along with their UMLS translations, constitute the reference lexicon for the evaluation.

|  | EN → FR | EN → DE |
|---|---|---|
| EVALUATION dataset | 126 → 163 | 90 → 104 |
| TRAINING dataset | 642 → 1953 | 584 → 1826 |

TABLE 4: Size of datasets

**The training dataset** contains source terms which could not be translated with the general-language dictionary or the UMLS but for which we could generate translations with our method. This generated translations were manually annotated by translators. These terms, along with their annotated translations, were used as training data for learning the ranking models and to tune the reliability values used in the RESO score. We used four classes for the annotation: *exact, acceptable, related* and *wrong*. An exact translation is a canonical translation like *cytoprotection → Zellschutz* (DE), *protection des cellules 'protection of the cells'* (FR). An acceptable translation is a variant of the canonical translation: *cytoprotection → protéger les cellules 'protect the cells', cytoprotecteur 'cytoprotective'*. A related translation is a translation which is only semantically related to the source term: *insecure → ohne Sicherheit 'without safety'*. All other translations are wrong translations. We computed inter-annotator agreement on a set of 100 randomly selected translations. We used the Kappa statistics (Carletta, 1996) and obtained a high agreement (0.77 for English to German translations and 0.71 for English to French).

## 4 Results

### 4.1 Related work

Generally, systems are compared using the *TopN* precision: the percentage of source terms with at least one exact translation among the *TopN* candidate translations. Compositional-translation methods tend to give better results when they are applied to general language texts rather than domain-specific texts. Indeed, it is easier to find translations of the components since they belong to the general language and large corpora are also easier to collect. **Working with general language texts,** Baldwin & Tanaka (2004) were able to generate candidate translations for 92% of their source terms and they report 43% (gold-standard) to 84% (silver standard) of correct translations on Top1. Corpus' size exceeds 80M words for each language. Cartoni (2009) works on the translation of prefixed Italian neologisms into French. He finds that between 42% and 94% of the generated neologisms occur more than five times on the Internet. Garera & Yarowsky (2008) obtain translations for 13% of the source words and the best precision is 39% for the Top10 candidate translations (for German and Swedish). **Regarding domain-specific translation,** Robitaille *et al.* (2006) collect translation pairs in an incremental manner. They start with a list of 9.6 pairs (on average) with a precision of 92% and end up with a final output of 19.6 pairs on average with a precision of 81%. Morin & Daille (2010) could generate candidate translations for 15% of their source terms and they report 88% of correct translations on Top1. The size of their corpus is 700k words per language. Weller *et al.* (2011) obtained correct English translations for 18% of their German compounds. Their corpus contains approximately 1.5M words per language.

## 4.2 Generation

We tested several combinations of linguistic resources. Table 5 only shows the results for the best combination. For English to French translation, the best results where obtained with all the combined resources, closely followed by the combination of the general language and the cognate dictionary. For English to German translation, the best results were obtained with the combination of the general language and the cognate dictionary. Morphologically related words and synonyms tend to increase the number of generated translations to the cost of translation accuracy.

|  | EN → FR | EN → DE |
|---|---|---|
| # source terms | 126 | 90 |
| # source terms with no translation | 40 (32%) | 34 (38%) |
| # source terms with at least one translation | 86 | 56 |
| # nb of translations / source term | 2.05 | 2.6 |
| # at least one reference translation (UMLS) | 68 (79%) | 40 (71%) |
| # at least one exact translation (translators or UMLS) | 81 (94%) | 51 (91%) |

TABLE 5: Results of generation

Regarding English to French translations, we were able to generate translations for 86 of the 126 English source terms (68%). Among these 86 source terms, 79% had the UMLS reference translation among their candidate translations. Regarding English to German translations, we were able to generate translations for 56 of the 90 English source terms (62%). Among these 56 source terms, 71% had a reference translation among their candidate translations. We noticed that the algorithm generated translations which were not in the UMLS lexicon but which were exact translations according to the translators. For example, the German reference translation for *mastectomy* is *mastektomie*. The system generated the reference translation *mastektomie* but also translations like *ablation der brust, abschnitt der brust, brustentfernun, entfernung der brust* which are all exact translations. Thus, if we take into account these translations, we find that 94% and 91% of the source terms had at least one exact translation for English to French and English to German respectively. Among all these correct translations 21% and 10% were fertile translations for English to French and English to German respectively.

There are several reasons for untranslated source terms. In 30% of the cases, silence is due to the coverage of the linguistic resources: some of the components could not be translated and the translation of the whole source term failed. Another 30% of target terms do not have a compositional meaning: *breastfeeding* → *allaitement* (FR), *stillen* (DE). The third reason is due to lexical variation (~ 20%), e.g. *radio+sensitivity* translates to *strahlen+toleranz* but *toleranz* 'tolerance' has a different meaning than *sensitivity*. There was also cases of semantic fertility (~ 13%).

Errors were mainly due to problems in word reordering when generating fertile translations, especially with German. Other errors were due to wrong translations in the cognate dictionary and translations which were inappropriate in the context, e.g. the translation of *gynae* to *frau*

'woman' in *gynaecomastia* → *Frau gegen Brust* 'women against breast'. If we look at the part of fertile translations in the incorrect translations, we find that they constitute half of the English to French incorrect translations and 80% of the English to German incorrect translations. We think it is due to the morphological type of the languages involved in the translation. As a matter of fact, fertile variants are more natural and more frequent in French than in German. English and German are Germanic languages with a tendency to build new words by agglutinating words or morphemes into one single word. Noun compounds such as *anthracycline-containing* or *Anthracyclin-enthaltende* are common in these two languages. Conversely, French is a Romance language which prefers to use phrases composed of two nouns and a preposition rather than a single-noun compound. For example, *anthracycline-containing* would be translated as *comprenant une anthracycline* 'containing an anthracycline'. There is no non-fertile equivalent in French (*\*anthracycline-contenant* would be ungrammatical). It is the same with the bound/free morpheme alternation. The term *cytotoxic* will be translated into German as *zytotoxisch* whereas in French it can be translated as *cytotoxique* or *toxique pour les cellules* 'toxic to the cells'.

## 4.3 Ranking

Table 6 indicates the precision on Top1, 2 and 3, i.e. the percentage of source terms which have at least one exact translation (found in the UMLS or according to the translators) on the TopN candidate translations.

|  | EN → FR | | | EN → DE | | | Average |
|---|---|---|---|---|---|---|---|
|  | Top1 | Top2 | Top3 | Top1 | Top2 | Top3 | Top1 |
| RANDOM | .83 | .88 | .93 | .80 | .88 | .88 | .815 |
| FREQ | .92 | .92 | .94 | .84 | .88 | .91 | .88 |
| POS | .88 | .93 | .94 | **.91** | **.91** | **.91** | .895 |
| CONT | .90 | .91 | .93 | .82 | .88 | .88 | .86 |
| RESO | .92 | .94 | .94 | .82 | .86 | .88 | .87 |
| COMBI | **.93** | **.94** | **.94** | .89 | .89 | .91 | **.91** |
| LTR ADARANK | .90 | .90 | .93 | .84 | .88 | .88 | .87 |
| LTR COORDINATE ASCEND | **.93** | **.94** | **.94** | .89 | .89 | .91 | **.91** |
| LTR LAMBDAMART | .86 | .91 | .93 | .88 | .91 | .91 | .87 |

TABLE 6: Results of ranking

We tested the ranking methods described in section 2.3: the scores FREQ, POS, CONT and RESO separately, a linear combination of these scores (COMBI) and three learning-to-rank algorithms (LTR). We also randomly ranked the translations to serve as a baseline. On average, the best precision on Top1 (91%) is obtained with the linear combination and Coordinate Ascend. All ranking methods perform better than the baseline. For English to French, the best rankings were obtained with the linear combination and Coordinate Ascend. For English to German, the best rankings were obtained with the Pos parameter alone, closely followed by the linear combination and Coordinate Ascend.

We expected learning-to-rank algorithms to perform much better than simple methods like part-of-speech probabilities or the linear combination of several scores. This might be due to the small size of our training dataset (approx. 600 ranked lists per language). We note that the context similarity score (Cont) is the least performing ranking method. Similarly, Garera and Yarowsky (2008) note only a small performance gain when they use context similarity. This might be due to the fact that context-based methods need the source and target words to be very frequent in the corpora to work properly. The lower quality of the German translations can be explained by the fact that the English-German corpus is much less comparable than the English-French corpus (0.45 vs. 0.71).

## Conclusion and perspectives

We have proposed a new compositional translation method for domain-specific bilingual lexicon extraction from comparable corpora. We obtain an average precision of 91% on the Top1 candidate translation. English-to-French translation performs slightly better than English-to-German translation, probably due to the morphological type of the languages and to the lower quality of the German data.

Future work includes the improvement of the identification of morphological variants. The morphological families extracted by the stemming algorithm are too broad for the purpose of translation. For example, the words *desirability* and *desiring* have the same stem but they are too distant semantically to be used to generate translation variants. We need to restrict the morphological families to a smaller set of morphological relations (e.g. noun → relational adjective). Furthermore, some work needs to be done on lexical variation: we used a dictionary of synonyms, but a thesaurus, which contains a large variety of semantic relations, may help us better in tackling lexical variation. Another improvement will be to use translation patterns to recompose the components into a target language term structure instead of using the Permutate and Concatenate functions.

Our investigation of the contribution of learning-to-rank algorithms to the problem of candidate translations ranking shows encouraging results and we should further pursue this line of research. Future experiments include: testing other learning-to-rank approaches (pair-wise, point-wise), increasing the number of predictive variables (e.g. source/target term frequency ratio, number of components...) and finding ways to increase the size of the training dataset at a lesser cost.

## Acknowledgment

The research leading to these results has received funding from the French National Research Agency under grant ANR-08-CORD-013 and from the European Community's Seventh Framework Programme (FP7/2007-2013) under Grant Agreement no 248005. We would like to thank the company Lingua et Machina (www.lingua-et-machina.com) for supporting this research as well as Clémence de Baudus and Kiril Isakov for their participation to the annotation task and Van Dang for his help with the RankLib software.

# References

Baker, M. (1996). Corpus-based translation studies: The challenges that lie ahead. In *Terminology, LSP and Translation: Studies in Language Engineering in Honour of Juan C. Sager*. Somers H., Amsterdam & Philadelphia, John Benjamins edition.

Baldwin, T., & Tanaka, T. (2004). Translation by Machine of Complex Nominals. *Proceedings of the ACL 2004 Workshop on Multiword expressions: Integrating Processing* (pp. 24-31). Barcelona, Spain.

Bo, L., & Gaussier, E. (2010). Improving Corpus Comparability for Bilingual Lexicon Extraction from Comparable Corpora. *Proceedings of the 23$^{th}$ International Conference on Computational Linguistics* (pp. 23-27). COLING 2010, Beijing, Chine.

Bodenreider, O. (2004). The Unified Medical Language System (UMLS): integrating biomedical terminology. *Nucleic Acids Research*, *32*, 267-270.

Bowker, L., & Pearson, J. (2002). *Working with Specialized Language: A Practical Guide to Using Corpora*. London/New York: Routledge.

Brown, P., Della Pietra, S., Della Pietra, V., & Mercer, R. (1993). The mathematics of statistical machine translation: parameter estimation. *Computational Linguistics*, *19*(2), 263-311.

Carletta, J. (1996). Assessing Agreement on Classification Tasks: The Kappa Statistic. *Computational Linguistics*, *22*(2), 249-254.

Cartoni, B. (2009). Lexical Morphology in Machine Translation: A Feasibility Study. *Proceedings of the 12th Conference of the European Chapter of the ACL* (pp. 130-138). EACL 2009, Athens, Greece.

Claveau, V., & Kijak, E. (2011). Morphological analysis of Biomedical Terminology with Analogy-Based Alignment. *Proceedings of Recent Advances in Natural Language Processing* (pp. 347-354). RANLP 2011, Hissar, Bulgaria.

Daille, B., & Morin, E. (2005). French-English terminology extraction from comparable corpora. *Proceedings of the 2nd International Joint Conference on Natural Language Processing*, Lecture Notes in Computer Science (Vol. 3651, pp. 707-718). Jeju Island, Korea: Springer.

Delpech, E. (2011). Evaluation of terminologies acquired from comparable corpora : an application perspective. *Proceedings of the 18$^{th}$ Nordic Conference of Computational Linguistics* (pp. 66-73). NODALIDA 2011, Riga, Latvia.

Delpech, E., Daille, B., Morin, E. & Lemaire, C. (2012). Identification of Fertile Translations in Medical Comparable Corpora : a Morpho-Compositional Approach. *Proceedings of the 10$^{th}$ Biennal Conference of the Association for Machine Translation in the Americas* (in press). AMTA 2012, San Diego, CA, USA.

Dunning, T. (1993). Accurate Methods for the Statistics of Surprise and Coincidence. *Computational Linguistics*, *19*(1), 61-74.

Fung, P. (1997). Finding Terminology Translations from Non-parallel Corpora. *Proceedings of the 5th Workshop on Very Large Corpora* (pp. 192-202), Hong Kong.

Fung, P., & Cheung, P. (2004). Mining Very-Non-Parallel Corpora: Parallel Sentence And Lexicon Extraction Via Bootstrapping And EM. *Proceedings of Empirical Methods in Natural Language Processing* (pp. 57-63). EMNLP 2004, Barcelona, Spain.

Garera N., & Yarowsky, D. (2008). Translating Compounds by Learning Component Gloss Translation Models via Multiple Languages. *Proceedings of the 3rd International Joint Conference on Natural Language Processing* (pp. 403-410). IJCNLP 2008, Hyderabad, India.

Grefenstette, G. (1999). The world wide web as a resource for example-based machine translation tasks. *Proceedings of the ASLIB Conference on Translating and the Computer*, *21*.

Hauer, B., & Kondrak, G. (2011). Clustering Semantically Equivalent Words into Cognate Sets in Multilingual Lists. *Proceedings of the 5th International Joint Conference on Natural Language Processing* (pp. 865–873). IJCNLP 2011, Chiang Mai, Thailand.

Keenan, E. L., & Faltz, L. M. (1985). *Boolean semantics for natural language*. Dordrecht, Holland: D. Reidel.

Lardilleux, A. (2008). A truly multilingual, high coverage, accurate, yet simple, sub-sentential alignment method. *The 8th conference of the Association for Machine Translation in the Americas* (pp. 125-132). AMTA 2008, Waikiki, Honolulu, USA.

Li, H., & Xu, J. (2007). AdaRank: A Boosting Algorithm for Information Retrieval. *Proceedings of the 30th annual international ACM SIGIR conference on Research and development in information retrieval* (pp. 391-398). SIGIR '07, Amsterdam, The Netherlands.

Li, H. (2011). Learning to Rank for Information Retrieval and Natural Language Processing. *Synthesis Lectures on Human Language Technology, Lecture 12*. Morgan & Claypool Publishers.

Liu, T.-Y. (2009). WWW 2009 Tutorial on learning to rank for information retrieval. *18th International World Wide Web Conference*. WWW 2009, Madrid, Spain.

Manning, C., Raghavan, P., & Schütze, H. (2008). *Introduction to Information Retrieval.* Cambridge University Press. New York, NY, USA.

Martinet, A. (1979). *Grammaire fonctionnelle du français.* Crédif/Dider. Paris, France.

Metzler, D., & Croft, W. B. (2000). Linear feature-based models for information retrieval. *Information Retrieval*, *10*(3), 257-274.

Morin, E., & Daille, B. (2010). Compositionality and lexical alignment of multi-word terms. *Language Resources and Evaluation (LRE)*, Multiword expression: hard going or plain sailing (Springer Netherlands., Vol. 44, pp. 79–95). Rayson, P. and Piao, S. and Sharoff, S. and Evert, S. and Villada Moirón.

Morin, E., Daille, B., Takeuchi, K., & Kageura, K. (2007). Bilingual terminology mining - using brain, not brawn comparable corpora. *Proceedings of the 45th Annual Meeting of the Association for Computational Linguistics* (pp. 664-671). ACL 2007, Prague, Czech Republic.

Namer, F., & Baud, R. (2007). Defining and relating biomedical terms: Towards a cross-language morphosemantics-based system. *International Journal of Medical Informatics*, *76*(2-3), 226-33.

Porter, M. F. (1980). An algorithm for suffix stripping. *Program*, *14*(3), 130-137.

Quasthoff, U., Richter, M., & Biemann, C. (2006). Corpus Portal for Search in Monolingual Corpora. *Proceedings of the fifth international conference on Language Resources and Evaluation* (pp. 1799-1802). LREC 2006, Genoa, Italy.

Rapp, R. (1995). Identifying Word Translations in Non-Parallel Texts. *Proceedings of the 33rd Conference of the Association for Computational Linguistics* (pp. 320-322). ACL'95, Boston, Massachussets, USA.

Rauf, S., & Schwenk, H. (2009). On the use of comparable corpora to improve SMT performance. *Proceedings of the 12th Conference of the European Chapter of the ACL* (pp. 16-23). EACL 2009, Athens, Greece.

Robitaille, X., Sasaki, X., Tonoike, M., Sato, S., & Utsuro, S. (2006). Compiling French-Japanese terminologies from the Web. *Proceedings of the 11th Conference of the European Chapter of the Association for Computational Linguistics* (pp. 225-232). EACL'06, Trento, Italy.

Tiedemann, J. (2009). News from OPUS - A Collection of Multilingual Parallel Corpora with Tools and Interfaces. *Recent Advances in Natural Language Processing*, John Benjamins (Vol. V, pp. 237-248). Amsterdam, Philadelphia: N. Nicolov and K. Bontcheva and G. Angelova and R. Mitkov.

Weller, M., Gojun, A., Heid, U., Daille, B., & Harastani, R. (2011). Simple methods for dealing with term variation and term alignment. *Proceedings of the 9th International Conference on Terminology and Artificial Intelligence* (pp. 87-93). TIA 2011, Paris, France.

Wu, Q., Burges, J. C., Svore, K., & Gao, J. (2010). Adapting Boosting for Information Retrieval Measures. *Journal of Information Retrieval*, *13*(3), 254 - 270.